\documentclass{article}

    \usepackage[preprint]{neurips_2020}

\usepackage[utf8]{inputenc} 
\usepackage[T1]{fontenc}    
\usepackage{hyperref}       
\usepackage{url}            
\usepackage{booktabs}       
\usepackage{amsfonts}       
\usepackage{nicefrac}       
\usepackage{microtype}      
\usepackage{amsmath}
\usepackage{caption}
\usepackage{subcaption}
\usepackage{graphicx}
\usepackage{amsmath}
\usepackage{caption}
\usepackage{subcaption}
\usepackage{wrapfig}
\newcommand{\bs}{\boldsymbol}
\def\be{\begin{equation}}
\def\ee{\end{equation}}
\def\bea{\begin{eqnarray}}
\def\eea{\end{eqnarray}}
\usepackage{xcolor}

\makeatletter
\def\@fnsymbol#1{\ensuremath{\ifcase#1\or \bs *\or \dagger\or \ddagger\or
   \mathsection\or \mathparagraph\or \|\or **\or \dagger\dagger
   \or \ddagger\ddagger \else\@ctrerr\fi}}
\makeatother

\setcitestyle{numbers,square,comma}

\title{ReZero is All You Need:\\ Fast Convergence at Large Depth}

\author{%
    \bf Thomas Bachlechner\thanks{Authors contributed equally, ordered by last name},\,  
    \bf Bodhisattwa Prasad Majumder\footnotemark[1],\,
    \bf Huanru Henry Mao\footnotemark[1],\,
    \\
    \bf Garrison W. Cottrell,\,
    \bf Julian McAuley
    \\
    UC San Diego \\
    \{\texttt{tbachlechner@physics},
    \texttt{bmajumde@eng},
    \texttt{hhmao@eng},\\
    \texttt{gary@eng},
    \texttt{jmcauley@eng}\}\texttt{.ucsd.edu} \\
}

\begin{document}

\maketitle

\begin{abstract}
Deep networks often suffer from vanishing or exploding gradients due to inefficient signal propagation, leading to long training times or convergence difficulties.
Various architecture designs, sophisticated residual-style networks, and initialization schemes have been shown to improve deep signal propagation.
Recently, Pennington \emph{et al.}~used free probability theory to show that dynamical isometry plays an integral role in efficient deep learning.
We show that the simplest architecture change of gating each residual connection using a single zero-initialized parameter satisfies initial dynamical isometry and outperforms more complex approaches.
Although much simpler than its predecessors, this gate enables training thousands of fully connected layers with fast convergence and better test performance for ResNets trained on CIFAR-10.
We apply this technique to language modeling and find that we can easily train 120-layer Transformers.
When applied to 12 layer Transformers, it converges 56\% faster on enwiki8.
\end{abstract}

\section{Introduction}

\begin{wrapfigure}[17]{r}{0.25\textwidth}
  \begin{center}
    \includegraphics[width=0.25\textwidth]{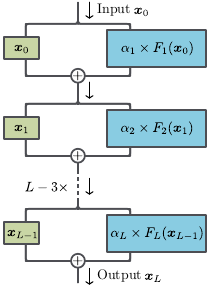}
  \end{center}
  \caption{ReZero\label{introfig}}
\end{wrapfigure}
Deep learning has enabled significant improvements in state-of-the-art performance across  domains \citep{lecun2015deep,he2016deep,klambauer2017self,radford2019language}.
The expressivity of neural networks typically grows exponentially with depth \citep{poole2016exponential}, enabling strong generalization performance, but often induces vanishing/exploding gradients and poor signal propagation through the model \citep{he2015delving}.
Practitioners have relied on residual \cite{he2016deep} connections along with complex gating mechanisms in highway networks \cite{srivastava2015highway}, careful initialization \citep{mishkin2015all,xiao2018dynamical,zhang2019fixup} and normalization such as BatchNorm \citep{ioffe2015batch} and LayerNorm \citep{ba2016layer} to mitigate this issue.

Recent theoretical work \cite{pennington2017resurrecting} applied free probability theory to randomly initialized networks and demonstrated that dynamical isometry is a key indicator of trainability.
Motivated by this theory, we study the simplest modification of deep networks that ensures initial dynamical isometry, which we call ReZero. 
ReZero
is a small addition to any  network that dynamically facilitates well-behaved gradients and arbitrarily deep signal propagation\footnote{Code for ReZero applied to various neural architectures: \url{https://github.com/majumderb/rezero}}. The idea is simple: ReZero initializes each layer to perform the identity operation. For each layer, we introduce a residual connection for the input signal $\bs x$ and one trainable parameter $\alpha$ that modulates the non-trivial transformation of a layer $F(\bs x)$,
\be
\bs x_{i+1} = \bs x_i + \alpha_i F(\bs x_i)\,,
\ee
where $\alpha=0$ at the beginning of training.
Initially the gradients for all parameters defining $F$ vanish, but dynamically evolve to suitable values during initial stages of training. We illustrate the architecture in Figure \ref{introfig}. 
ReZero provides several benefits:

{\bf Widely Applicable:} Unlike more complex schemes, ReZero is simple and architecture agnostic, making its implementation widely applicable to any residual-style architectures without much tuning and only a few lines of code.

{\bf Deeper learning:} While simpler than existing approaches \cite{srivastava2015highway,zhang2019fixup}, ReZero effectively propagates signals through deep networks, which allows for learning in otherwise untrainable networks. ReZero successfully trains 10,000 layers of fully-connected networks, and we are the first to train Transformers over 100 layers without learning rate warm-up, LayerNorm \cite{ba2016layer} or auxiliary losses \citep{al2019character}.

{\bf Faster convergence:} We observe accelerated convergence in ReZero networks compared to regular residual networks with normalization. When ReZero is applied to Transformers, we converge 56\% faster than the vanilla Transformer to reach 1.2 BPB on the enwiki8 language modeling benchmark. When applied to ResNets, we obtain 32\% speed up to reach 85\% accuracy on CIFAR 10.

\section{Background and related work}
\begin{table}
  \caption{Various forms of normalization and residual connections. $F$ represents the transformation of an arbitrary layer and ``Norm'' is a normalization  (e.g.~LayerNorm or BatchNorm).}
  \label{tab:resarchs}
  \centering
  \begin{tabular}{ll}
    \toprule
    (1) Deep Network & $\bs x_{i+1} = F(\bs x_i)$ \\
    (2) Residual Network     & $\bs x_{i+1} = \bs x_i +F(\bs x_i)$ \\
    (3) Deep Network + Norm    & $\bs x_{i+1} = \text{Norm}\big(F(\bs x_i)\big)   $  \\
    (4) Residual Network + Pre-Norm    & $\bs x_{i+1} =  \bs x_i + F(\text{Norm}\big(\bs x_i\big))  $  \\
    (5) Residual Network + Post-Norm     & $\bs x_{i+1} =   \text{Norm}\big(\bs x_i + F(\bs x_i)\big)   $  \\
    (6) {\textbf{ReZero}}& $\bs x_{i+1} = \bs x_i + \alpha_i  F(\bs x_i)$\\
    \bottomrule
  \end{tabular}
\end{table}

Networks with a depth of $L$ layers and width $w$ often have an expressive power that scales exponentially in depth, but not in width \citep{montufar2014number,poole2016exponential}.
Large depth often comes with difficulty in training via gradient-based methods.
During training of a deep model, a signal in the training data has to propagate forward from the input to the output layer, and subsequently, the cost function gradients have to propagate backwards in order to provide a meaningful weight update.
If the magnitude of a perturbation is changed by a factor $r$ in each layer, both signals and gradients vanish or explode at a rate of $r^L$, rendering many deep networks untrainable in practice.

To be specific, consider a deep network that propagates an input signal $\bs x_0$ of width $w$ through $L$ layers that perform the non-trivial, but width preserving functions $F[{\cal W}_i]: \mathbb R^w\rightarrow \mathbb R^w$, where ${\cal W}_i$ denotes all parameters at layer $i=1,\dots,L$. The signal propagates through the network according to
\be
\bs x_{i+1} = F[{\cal W}_i](\bs x_i)\,.\label{deepnetwork}
\ee
There have been many attempts to improve signal propagation through deep networks and they often fall into one of three categories---initialization schemes, normalization, and residual connections. We show some of the popular ways to combine residual networks with normalization in Table \ref{tab:resarchs}.

\subsection{Careful initialization}
The dynamics of signal propagation in randomly initialized deep and wide neural networks have been formalized via mean field theory \citep{pennington2017resurrecting,xiao2018dynamical,pennington2018emergence}. 
For some deep neural networks, including fully connected and convolutional architectures, the cosine distance of two distinct signals, $\bs x_i\cdot \bs x'_i/(\|\bs x_i\|\|\bs x'_i\|)$, approaches a fixed point that either vanishes or approaches unity at large depths.
If this fixed point is $1$ the behavior of the network is stable and every input is mapped to the same output, leading to vanishing weight updates. If this fixed point is $0$ the behavior of the network is chaotic and even similar inputs are mapped to very different outputs, leading to exploding weight updates. To understand whether a network is in a stable or chaotic phase we consider the input-output Jacobian
\be
\bs J_{\text{io}}\equiv \frac{\partial \bs x_L}{ \partial \bs x_0}\,.
\ee
The mean squared singular values $\chi$ of this matrix determine the  growth/decay of an average input signal perturbation as it propagates through the network. The network exhibits a boundary between the ordered and the chaotic phase, the edge of chaos at $\chi = 1$. Training proceeds efficiently at the edge of chaos.
This behavior was recognized in \citep{glorot2010understanding,he2015delving}, which motivated a re-scaling of the weights such that $\chi \approx 1$ and on average signal strengths are neither enhanced or attenuated.

Pennigton \textit{et al.}~\citep{pennington2017resurrecting,pennington2018emergence} recognized that a unit mean squared average of the input-output Jacobian is insufficient to guarantee trainability. For example, if the singular vectors of $\bs J_{\text{io}}$ corresponding to very large/small singular values align well with the perturbations in the data, training will still be inefficient. They proposed the stronger condition of {\it dynamical isometry} \citep{saxe2013exact}, which requires that all singular values of $\bs J_{\text{io}}$ are close to one. This means that all perturbations of the input signal propagate through the network equally well. The ReLU activation function maps to zero for some perturbations of the input signal, and it is therefore intuitive that deep networks with ReLU activations cannot possibly satisfy dynamical isometry, as was rigorously established in \citep{pennington2017resurrecting}. For some activation functions and network architectures, elaborate initialization schemes allow the network to satisfy dynamical isometry at initialization, which significantly improves training dynamics  \citep{schoenholz2016deep,poole2016exponential,yang2017mean,gilboa2019dynamical}. 

\subsection{Normalization}
An alternative approach to improve the trainability of deep networks is to incorporate layers that explicitly provide normalization.
Many normalization modules have been proposed, with the two most popular ones being BatchNorm \citep{ioffe2015batch} and LayerNorm \citep{ba2016layer}.
In general, normalization aims to ensure that initially, signals have zero mean and unit variance as they propagate through a network, reducing \textit{covariate shift} \citep{ioffe2015batch}.

Normalization methods have shown success in accelerating the training of deep networks, but they do incur a computational cost to the network and pose additional hyperparameters to tune (e.g.~where to place the normalization). In contrast to normalization methods, our proposed method is simple and cheap to implement. ReZero alone is sufficient to train deeper networks, even in the absence of various norms.
Although ReZero makes normalization superfluous for convergence, we have found the regularizing effect of BatchNorm to be complementary to our approach.

\subsection{Residual connections}
The identity mappings introduced for ResNet in \citep{he2016deep} enabled a deep residual learning framework in the context of convolutional networks for image recognition that significantly increased the trainable depth. In \citep{he2016identity} it was recognized that identity (pre-activation) residual connections allow for improved signal propagation. Residual connections in ResNets allowed for training of extremely deep networks, but the same has not been the case for Transformer architectures. Deep Transformer architectures 
have thus far
required extreme compute budgets or auxiliary losses. 

Careful initialization has been employed in conjunction with residual connections. It has been proposed to initialize residual blocks around zero in order to facilitate better signal propagation \cite{srivastava2015highway,he2016identity,goyal2017accurate,hardt2016identity,he2019bag,zhang2019fixup}. 
Concurrently with our work SkipInit \citep{de2020batch}, an alternative to the BatchNorm, was proposed  for ResNet architectures that is  similar to ReZero.
The authors find that in deep ResNets without BatchNorm, a scalar multiplier is needed to ensure convergence.
We arrive at a similar conclusion for the specific case considered in \citep{de2020batch}, and study more generally signal propagation in deeper networks across multiple architectures and beyond BatchNorm. 

\section{ReZero}
We propose \textbf{ReZero} (\textbf{re}sidual with \textbf{zero} initialization), a simple change to the architecture of deep residual networks that facilitates dynamical isometry and enables the efficient training of extremely deep networks. Rather than propagating the signal through each of the non-trivial functions $F[{\cal W}_i]$ at initialization, we add a skip connection and rescale the function by $L$ learnable parameters $\alpha_i$ (which we call \textit{residual weights}) that are initialized to zero. The signal now propagates according to 
\be
\bs x_{i+1} = \bs x_i + \alpha_i F[{\cal W}_i](\bs x_i)\,.\label{deepnetwork2}
\ee
At initialization the network  represents the identity function and it trivially satisfies dynamical isometry.
We demonstrate below for a toy model that this architecture can exponentially accelerate training.
The architecture modification allows for the training of deep networks even when the individual layers' Jacobian has vanishing singular values, as is the case for ReLU activation functions or self-attention \citep{vaswani2017attention}.

\subsection{A toy example}
\begin{wrapfigure}[28]{r}{0.4\textwidth}
  \centering
  \includegraphics[width=0.4\textwidth]{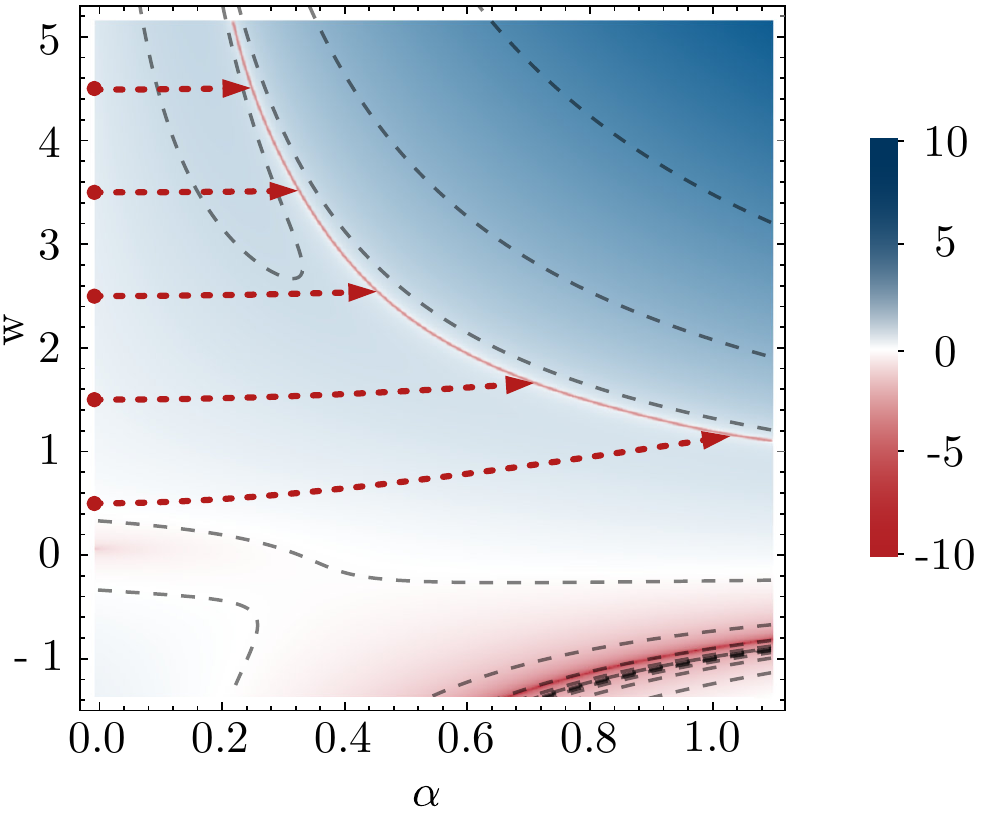}
  \caption{\footnotesize Contour plot of the log gradient norm, $\text{log}\lVert{\partial {\cal C}}\lVert_2$, over the network weight $w$ and the residual weight $\alpha$  during the training of the linear function $x_{L=5} = 50\times x_0$ via gradient descent using a training set of $x_0=\{1., 1.1, \dots, 1.8\}$. Gradient descent trajectories initialized at $\alpha = 0$ are shown in red for five different initial $w$'s. The trajectory dynamics avoid the poorly conditioned regions around $\alpha\approx 1$ and converge to the solution $\alpha w\approx 1.2$. 
  \label{fig:toymodel}}
\end{wrapfigure}

To illustrate how the ReZero connection accelerates training let us consider the toy model of a deep neural network described by $L$ single-neuron hidden layers that have no bias and all share the same weight $w$ and $\alpha_i=\alpha ~\forall i$. The network then simply maps an input $x_0$ to the output
\be
x_L = (1+\alpha w)^L x_0\,.
\ee
Fixing the parameter $\alpha= 1$ would represent a toy model for a fully connected residual network, while initializing $\alpha=0$ and treating $\alpha$ as a learned parameter corresponds to a ReZero network.
The input-output Jacobian is given by $J_{\text{io}}= (1+\alpha w)^L$, indicating that for initialization with $w\approx 1$ and $\alpha = 1$ the output signal of a deep (e.g.,~$L\gg 1$) network is extremely sensitive to any small perturbations of the input, while with $\alpha = 0$ the input signal magnitude is preserved.  While this example is too simple to exhibit an order/chaos phase transition, it does accurately model the vanishing and exploding gradient problem familiar in deep networks. Assuming a learning rate $\lambda$ and a cost function $\cal C$, gradient descent updates the weights $w$ according to
\be
w \leftarrow w-\lambda L\alpha x_0 (1+\alpha w)^{L-1}\partial_x {\cal C}(x)|_{x=x_L}\,.
\label{eq:gradient_toy}\ee

For $\alpha = 1$, convergence of gradient descent with an initial weight $w\approx 1$ requires steps no larger than 1, and hence a learning rate that is exponentially small in depth $L$
\be
\lambda \propto L^{-1}(1+ w)^{-(L-1)}\,,
\ee
where we only retained the parametric dependence on $w$ and $L$.
For $w\gg 1$ the gradients in Equation \ref{eq:gradient_toy} explode, while for $w\approx -1$ the gradients vanish. Initializing $\alpha =0$ solves both of these problems: assuming a sufficiently well-conditioned cost function, the first step of gradient descent will update the residual weights $\alpha$ to a value that avoids large outputs and keeps the parameter trajectory within a well-conditioned region while retaining the expressive power of the network. The first non-trivial steps of the residual weight updates are given by
\be
\alpha \leftarrow -\lambda Lw x_0\partial_x {\cal C}(x)|_{x=x_L}\,,
\ee

and gradient descent will converge with a learning rate that is polynomial in the depth $L$ of the network. In this simple example, the ReZero connection, therefore, allows for convergence with dramatically fewer optimization steps compared to a vanilla residual network.
We illustrate the training dynamics and gradients in Figure \ref{fig:toymodel}.

\section{Training deep fully-connected networks faster}
\label{sec:fc}

We now study the effect of ReZero on deep ReLU networks, and compare it with some of the approaches that facilitate deep learning listed in the rows of Table \ref{tab:resarchs}. Specifically, we will compare a vanilla deep fully connected network (FC, row 1), a deep network with residual connections (FC+Res, row 2), a deep network with LayerNorm (FC+Norm, row 3), and finally our proposed ReZero (row 6). We choose the initial weights $\bs W_i$ to be normally distributed with variances optimal for training \citep{he2015delving,yang2017mean}, e.g.,~$\sigma^2_{W}=2/w$ for all but the vanilla residual network where $\sigma^2_{W}\approx 0.25/w$.

As a sample toy task, we train four different 32-layer network architectures on the CIFAR-10 data set for supervised image classification. We are only interested in the training dynamics and investigate how many iterations it takes for the model to fit the data.

\begin{wrapfigure}[17]{r}{0.5\textwidth}
  \centering
  \includegraphics[width=0.5\textwidth]{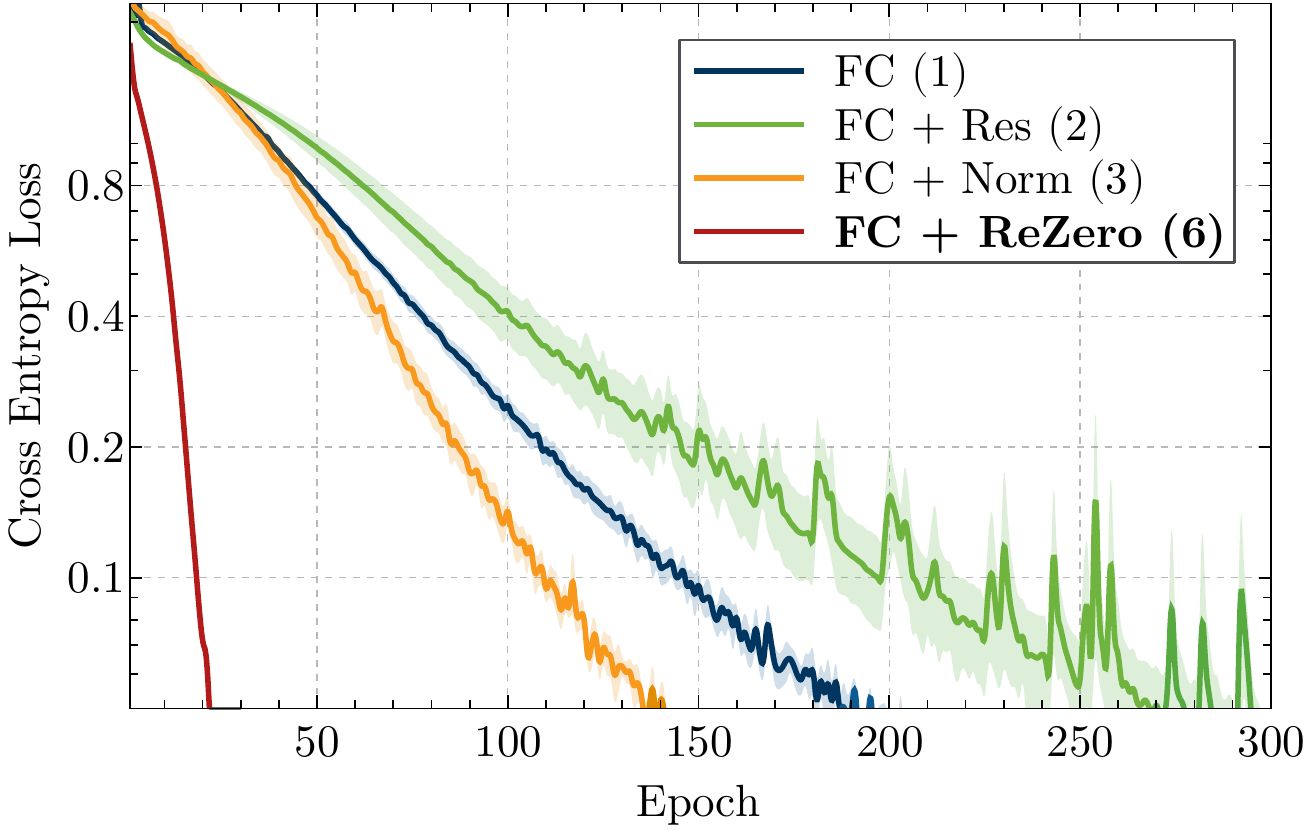}
  \caption{\footnotesize Cross entropy loss during training of four variants of $32$ layer fully-connected networks with width $256$ and ReLU activations.
  Numbers in parentheses refer to the architectures in the corresponding rows of Table \ref{tab:resarchs}.  We average over five runs each and show $1\sigma$ error bands. We train using Adagrad \citep{duchi2011adaptive} with learning rate $0.01$. \label{fig:fctraining}}
\end{wrapfigure}

We show the evolution of the training loss in Figure \ref{fig:fctraining}. In our simple experiment, the ReZero architecture converges to fit the training data between 7 and 15 times faster than other techniques. Note that without an additional normalization layer the residual connection decreases convergence speed compared to a plain fully connected network. We speculate that this is because at initialization the variance of the signal is not independent of depth, see \citep{yang2017mean}.

With increasing depth, the advantages of the ReZero architecture become more apparent. To verify that this architecture ensures trainability to large depth we successfully trained fully connected ReZero networks with up to $10,000$ layers on a laptop with one GPU\footnote{To train at these extreme depths we used the Adagrad optimizer with a learning rate of $0.003$.} to overfit the training set.

\section{Training Convolutional ResNets faster}\label{sec:resnet}
Residual connections enabled the first widely used feed-forward networks for image recognition with hundreds of layers \cite{srivastava2015highway,he2016deep}. It was quickly realized that applying the residual connection after the activation function (in PreAct-ResNets \cite{he2016identity}, see Table \ref{tab:resarchs}), as well as initializing the network closer to the identity mapping (in \cite{savarese2016learning,goyal2017accurate,hardt2016identity,he2019bag,zhang2019fixup,de2020batch}) leads to improved performance. ReZero combines the benefits of both approaches and is the simplest implementation in this sequence of papers.

In the previous section, we saw how ReZero connections enable the training of networks with thousands of layers
that would otherwise be untrainable. In this section, we apply ReZero connections to deep convolutional residual networks for image recognition \citep{he2016deep}. While these networks are trainable without modification, we observe that ReZero accelerates training and improves 
accuracies.

In order to compare different methods (ResNet \cite{he2016deep} modified by Gated ResNet \cite{srivastava2015highway,savarese2016learning}, zero $\gamma$ \cite{goyal2017accurate,hardt2016identity}, FixUp \cite{zhang2019fixup}, ReZero and Pre-Act ResNet \cite{he2016identity} modified with ReZero) to improve deep signal propagation, we trained various versions of residual networks on the CIFAR-10 image classification dataset, each with identical hyperparameters. We discuss the architectures and hyperparameters in detail in Appendices \ref{sec:residualgates} and \ref{sec:cifarhyperparameters}. In Table \ref{tab:resnetablations} we present results for the validation error, the number of epochs to reach $80\%$ accuracy, and loss on the training data. ReZero performs better than the other methods for ensuring deep signal propagation: it accelerates training as much as FixUp, but retains the regularizing properties of the BatchNorm layer. Gated ResNets initially train very fast, but perform significantly worse than ReZero on test data.

\begin{table*}
  \small
  \centering
  \caption{Comparison of ResNet variants on CIFAR-10, see Appendix \ref{sec:cifarhyperparameters} for implementation details. The uncertainties correspond to standard error.}
      \label{tab:resnetablations}
      \begin{tabular}{lrrrr}
        \toprule
        Model & Val. Error [$\%$] & Change &  Epochs to 80$\%$ Acc. & Train Loss $\times$1000 \\
        \midrule
        ResNet-56 \cite{he2016deep} & 6.27~$\pm$~0.06 & --&20~$\pm$~1 &5.9~$\pm$~0.1 \\
        \hspace{10pt} + Gated ResNet \cite{srivastava2015highway,savarese2016learning} &6.80~$\pm$~0.09 & $+$~0.53 &9~$\pm$~2 &4.6~$\pm$~0.3 \\
        \hspace{10pt} + zero $\gamma$ \cite{goyal2017accurate,hardt2016identity}  & 7.84~$\pm$~0.05 & $+$~1.57 &39~$\pm$~4 &31.2~$\pm$~0.5 \\
        \hspace{10pt} + FixUp \cite{zhang2019fixup} & 7.26~$\pm$~0.10 & $+$~0.99 &13~$\pm$~1 &4.6~$\pm$~0.2 \\
        \hspace{10pt} + {\textbf{ReZero}} & 6.58~$\pm$~0.07 & $+$~0.31 &15~$\pm$~2 &4.5~$\pm$~0.3  \\
        \midrule
        ResNet-110  \cite{he2016deep}  & 6.24~$\pm$~0.29 & --&23~$\pm$~4 &4.0~$\pm$~0.1 \\
        \hspace{10pt} + Gated ResNet \cite{srivastava2015highway,savarese2016learning} & 6.71~$\pm$~0.05 & $+$~0.47 &10~$\pm$~2 &2.8~$\pm$~0.2 \\
        \hspace{10pt} + zero $\gamma$ \cite{goyal2017accurate,hardt2016identity} & 7.49~$\pm$~0.07 & $+$~1.25 &36~$\pm$~5 &18.5~$\pm$~0.9 \\
        \hspace{10pt} + FixUp \cite{zhang2019fixup} & 7.10~$\pm$~0.22 & $+$~0.86 &15~$\pm$~1 &3.3~$\pm$~0.5 \\
        \hspace{10pt} + {\textbf{ReZero}} & 5.93~$\pm$~0.12 & $-$~0.31  &14~$\pm$~1 &2.6~$\pm$~0.1 \\
        \midrule
        Pre-activation ResNet-18 \cite{he2016identity} & 6.38~$\pm$~0.01 & -- &26~$\pm$~2 &4.1~$\pm$~0.3  \\
        \hspace{10pt} + {\textbf{ReZero}} & 5.43~$\pm$~0.06 & $-$~0.95 &12~$\pm$~1 &1.9~$\pm$~0.3 \\
        \midrule
        Pre-activation ResNet-50 \cite{he2016identity} & 5.37~$\pm$~0.02  & -- &26~$\pm$~3 &2.6~$\pm$~0.1  \\
        \hspace{10pt} + {\textbf{ReZero}} &  4.80$~\pm$ 0.08 & $-$~0.57 &17~$\pm$~1 &2.2~$\pm$~0.1  \\
        \bottomrule
      \end{tabular}
\end{table*}
In order to demonstrate the accelerated training of ReZero networks, we implemented superconvergence \cite{smith2019super} in a Pre-activation ResNet-18 with ReZero connections. The phenomenon of superconvergence uses one cycle of an increasing and decreasing learning rate, in which a large maximum learning rate serves as a regularizer. This yields very fast convergence for some networks. We find that the training duration to achieve $94\%$ accuracy decreases from the 60 epochs for the baseline\footnote{Our implementation was inspired by the codes from fast.ai available at \url{https://github.com/fastai/imagenet-fast/tree/master/cifar10}. We replicated this model and added ReZero.
It was important to have a small, constant learning rate for the residual weights, otherwise the ReZero model diverges easily.} model to 45 epochs for a ReZero model.

\section{Training deeper Transformers faster}
In this section, we study the signal propagation and application of ReZero to the Transformer architecture \citep{vaswani2017attention}. Transformers gained significant popularity and success both in supervised and unsupervised NLP tasks \citep{DBLP:conf/naacl/DevlinCLT19, al2019character}.
Transformers are built by stacking modules that first perform self-attention, then a point-wise feed-forward transformation. 

\begin{wrapfigure}[18]{r}{0.45\textwidth}
  \centering
  \includegraphics[width=0.45\textwidth]{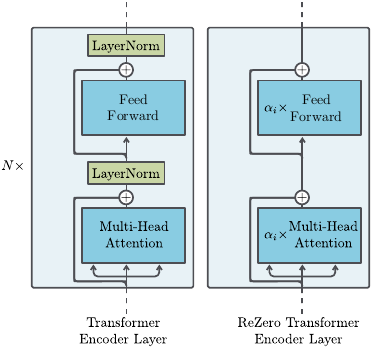}
  \caption{\footnotesize ReZero for Transformers\label{fig:rezerotransformer}}
\end{wrapfigure}

The original Transformer \citep{vaswani2017attention} implementation can be seen as a residual network with post-normalization (row 5 in Table \ref{tab:resarchs}).
Inside a Transformer module the output of each sublayer is added via a residual connection and then normalized by LayerNorm,
\be
\label{eq:tx}
\bs x_{i + 1} = \text{LayerNorm}\left(\bs x_i + \text{sublayer}(\bs x_i)\right)\, 
\ee
where $\text{sublayer} \in \{\text{self-attention}, \text{feed-forward}\}$, as illustrated in the left panel of Figure \ref{fig:rezerotransformer}.

\subsection{Signal propagation in Transformers}
\label{sec:sigtransformers}

\begin{figure}
  \centering
  \includegraphics[width=.9\textwidth]{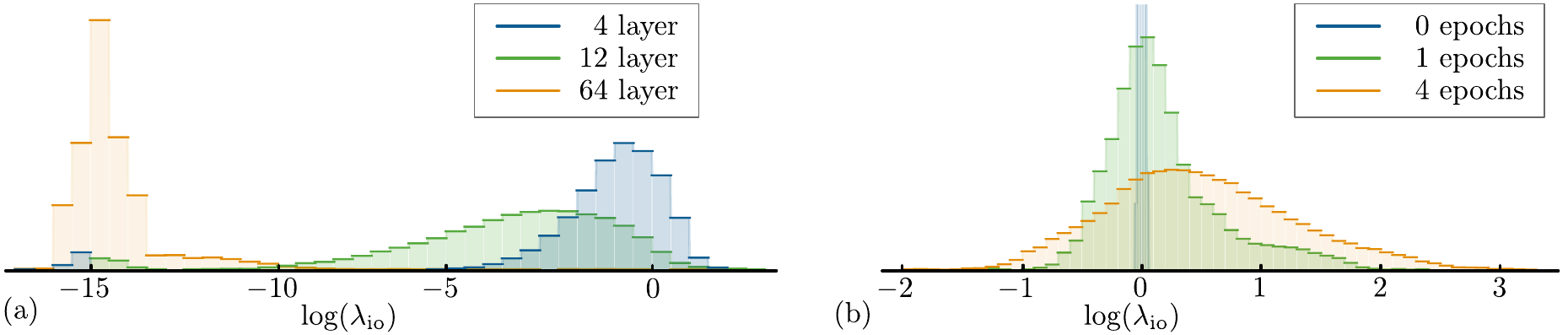}
  \caption{\footnotesize Histograms of log singular values ($\log(\lambda_{\text{io}})$)
  for the input-output Jacobian: (a) Transformer encoder at initialization; depths 4, 12 and 64 layers. (b) ReZero Transformer encoder  with 64 layers before/during training. Deep ReZero Transformers remain much closer to dynamical isometry, where mean singular value $\lambda_{\text{io}}\approx 1$. \label{fig:histograms}}
\end{figure}

Two crucial components relevant to the signal propagation in the original Transformer layers include LayerNorm \citep{ba2016layer} and (multi-head) self attention \citep{vaswani2017attention}.
Neither component by itself or in conjunction with a vanilla residual connection can satisfy dynamical isometry for all input signals, as we show with a theoretical argument in Appendix \ref{sec:singval}. 
We verify these claims in practice by evaluating the change of the attention output under an infinitesimal variation of each input element, which yields the input-output Jacobian.
We show the input-output Jacobian for Transformer encoder layers of various depths with Xavier uniform initialized weights in Figure \ref{fig:histograms}a. While shallow Transformers exhibit a singular value distribution peaked around unity, we clearly observe that the Jacobian of deeper Transformers has a large number of singular values that vanish to machine precision. While the distribution varies depending on the details of the initialization scheme, the qualitative statement holds more broadly. These results are consistent with the common observation that deep Transformer networks are extremely challenging to train.

In order to facilitate deep signal propagation we apply ReZero by replacing LayerNorm and re-scaling the self-attention block. Specifically, this modifies equation (\ref{eq:tx}) to
\be
\bs x_{i + 1} = \bs x_i + \alpha_i \text{sublayer}(\bs x_i)\,,
\ee
where $\alpha_i$ is the learned residual weight parameter as in the right panel of Figure \ref{fig:rezerotransformer}.
We share the same $\alpha_i$ parameter for a pair of multi-head self-attention and feed-forward network within a Transformer layer.
At initialization, $\alpha_i=0$, which allows for unimpeded signal propagation: All singular values of the input-output Jacobian are 1 and the model trivially satisfies dynamical isometry. To verify that the model remains close to dynamical isometry throughout training and for larger $\alpha_i$, we show a histogram of the Jacobian singular values during the training of a $64$ layer Transformer decoder language model on WikiText-2 \citep{DBLP:conf/iclr/MerityX0S17} in Figure \ref{fig:histograms}b. During training the weight of the residual connection gradually increases, allowing the Transformer to model extremely complex functions while maintaining signal propagation properties close to dynamical isometry.

\subsection{Convergence speed}
\label{sec:conveff}

We select language modeling on \texttt{enwiki8} \citep{mahoney2009} as a benchmark because strong language models are a good indicator of downstream NLP task performance \citep{radford2019language}.
Our aim in these experiments is to measure the convergence speed of each method by measuring the number of iterations it takes for a 12 layer Transformer to reach 1.2 bits per byte (BPB) on \texttt{enwiki8}.

Since the introduction of Transformers \citep{vaswani2017attention}, there have been several competing placements of the LayerNorm \cite{DBLP:journals/corr/abs-1910-05895, DBLP:journals/corr/abs-1910-10683} within the Transformer to achieve better convergence \citep{radford2019language,xiong2020layer}.
We experiment with 3 Transformer normalization methods and compare against the ReZero Transformer. The \textit{Post-Norm} (Row 5 in Table \ref{tab:resarchs}) method is equivalent to the vanilla Transformer in \citep{vaswani2017attention}, the \textit{Pre-Norm} (Row 4 in Table \ref{tab:resarchs}) method was recently introduced in \citep{xiong2020layer} and the \textit{GPT2-Norm} ($\bs x_{i+1} = \bs x_i + \text{Norm}(F(\bs x_i))$) was used in the training of GPT2 \citep{radford2019language}, which has successfully trained Transformers up to 48 layers. Finally, we experiment with our proposed ReZero method with $\alpha$ initialized to either zero or one. The hyperparameters are in Appendix  \ref{sec:appendix-tx1}.

Our results (Table \ref{tab:conveff}) show that \textit{Post-Norm} diverges during training while all other models are able to converge.
This is not surprising as the original Transformer implementation required a learning rate warm-up likely to overcome its poor initial signal propagation, as confirmed in \citep{xiong2020layer}.
To verify this, we re-ran the \textit{Post-Norm} setup with 100 steps of learning rate warm-up and find that the model is able to converge to 1.2 BPB in 13,690 iterations.
Under this setting, we compared other LayerNorm placement schemes against \textit{Post-Norm}. We find that the other placements led to initially faster convergence, but ultimately \textit{Post-Norm} catches up in performance, resulting in relatively slower convergence for \textit{Pre-Norm} and \textit{GPT2-Norm}.
However, other LayerNorm placements have an advantage over \textit{Post-Norm} in that they do not require learning rate warm-up, thus have fewer hyperparameters to tune.
ReZero with $\alpha = 1$ does not show an improvement over the vanilla Transformer, indicating the importance of initializing $\alpha = 0$.
With our proposed initialization of $\alpha = 0$, ReZero converges 56\% faster than the vanilla Transformer.

\begin{table}
\parbox{.45\linewidth}{
  \small
  \centering
  \caption{Comparison of various 12 layer Transformers normalization variants against ReZero and the training iterations required to reach 1.2 BPB on \texttt{enwiki8} validation set.}
      \label{tab:conveff}
      \begin{tabular}{lrr}
        \toprule
        Model & Iterations & Speedup\\
        \midrule
        Post-Norm \citep{vaswani2017attention} & Diverged & -\\
        \hspace{10pt} + Warm-up & 13,690 & 1$\times$\\
        Pre-Norm & 17,765 & 0.77$\times$\\
        GPT2-Norm \citep{radford2019language} & 21,187 &  0.65$\times$\\
       ReZero $\alpha  =  1$ & 14,506 & {0.94$\times$} \\
       {\bf ReZero} $\bs \alpha \bs = \bs 0$ & \textbf{8,800} & \textbf{1.56$\times$}\\
        \bottomrule
      \end{tabular}
  }
  \hspace{10pt}
  \parbox{.51\linewidth}{
  \small
  \caption{Comparison of Transformers (TX) on the \texttt{enwiki8} test set. Char-TX refers to the Character Transformer  \citep{al2019character} and uses additional auxiliary losses to achieve its performance.}
  \setlength{\tabcolsep}{4pt}
      \label{tab:deepformer}
      \begin{tabular}{lrrr}
        \toprule
        Model & Layers & Parameters & BPB\\
        \midrule
        Char-TX \citep{al2019character} & 12 & 41M & 1.11 \\
        TX + Warm-up & 12 & 38M & 1.17 \\
        TX + ReZero $\alpha=1$ & 12 & 34M & 1.17 \\
        TX + ReZero $\alpha=0$ & 12 & 34M & 1.17 \\
        \midrule
        Char-TX \citep{al2019character} & 64 & 219M & 1.06 \\
        TX & 64 & 51M & Diverged \\
        TX + Warm-up & 64 & 51M & Diverged \\
        TX + ReZero $\alpha=1$ & 64 & 51M & Diverged \\
        TX + ReZero $\alpha=0$ & 64 & 51M & 1.11 \\
        \midrule
        TX + ReZero & 128 & 101M & 1.08 \\
        \bottomrule
      \end{tabular}
      
  \setlength{\tabcolsep}{6pt}
  }
\end{table}

\subsection{Deeper Transformers}\label{sec:deeper-transformers}
Deeper Transformers require significantly more compute to train, with 78 layer Transformers requiring a cluster of 256 GPUs \citep{tnlg2020}.
This cost comes from an increase in memory requirements and poor signal propagation.
The Character Transformer \citep{al2019character} mitigated this issue by having intermediate layers predict the target objective as an auxiliary loss, thus circumventing vanishing gradients.
In this section, we extend our 12 layer ReZero Transformer from Section \ref{sec:conveff} to 64 and 128 layers and compare against the vanilla Transformer (\textit{Post-Norm}) and the Character Transformer.
Our results (Table \ref{tab:deepformer}) indicate that a 12 layer ReZero Transformer attains the same BPB as a regular Transformer after convergence, which shows that we do not lose any representational expressivity in our model by replacing LayerNorm with ReZero.
We find that trying to train deep vanilla Transformers leads to convergence difficulties.
When scaled to 64 layers, the vanilla Transformer fails to converge even with a warm-up schedule.
A ReZero Transformer with initialization of $\alpha=1$ diverges, supporting our theoretically motivated initialization at $\alpha=0$.
The deeper ReZero Transformers are able to attain better performance than the shallower Transformers.

We also display results from Character Transformer \citep{al2019character}, which had a similar setup, but required more parameters and used intermediate and other auxiliary losses to achieve their performance.
In contrast, our 128 layer Transformer achieves similar performance and learns effectively without any intermediate losses.
We did not tune our hyperparameters (Appendix \ref{sec:appendix-tx2}) and our models can potentially achieve better results with stronger regularization.

\begin{wrapfigure}[17]{r}{0.5\textwidth}
  \begin{center}
    \includegraphics[width=0.5\textwidth]{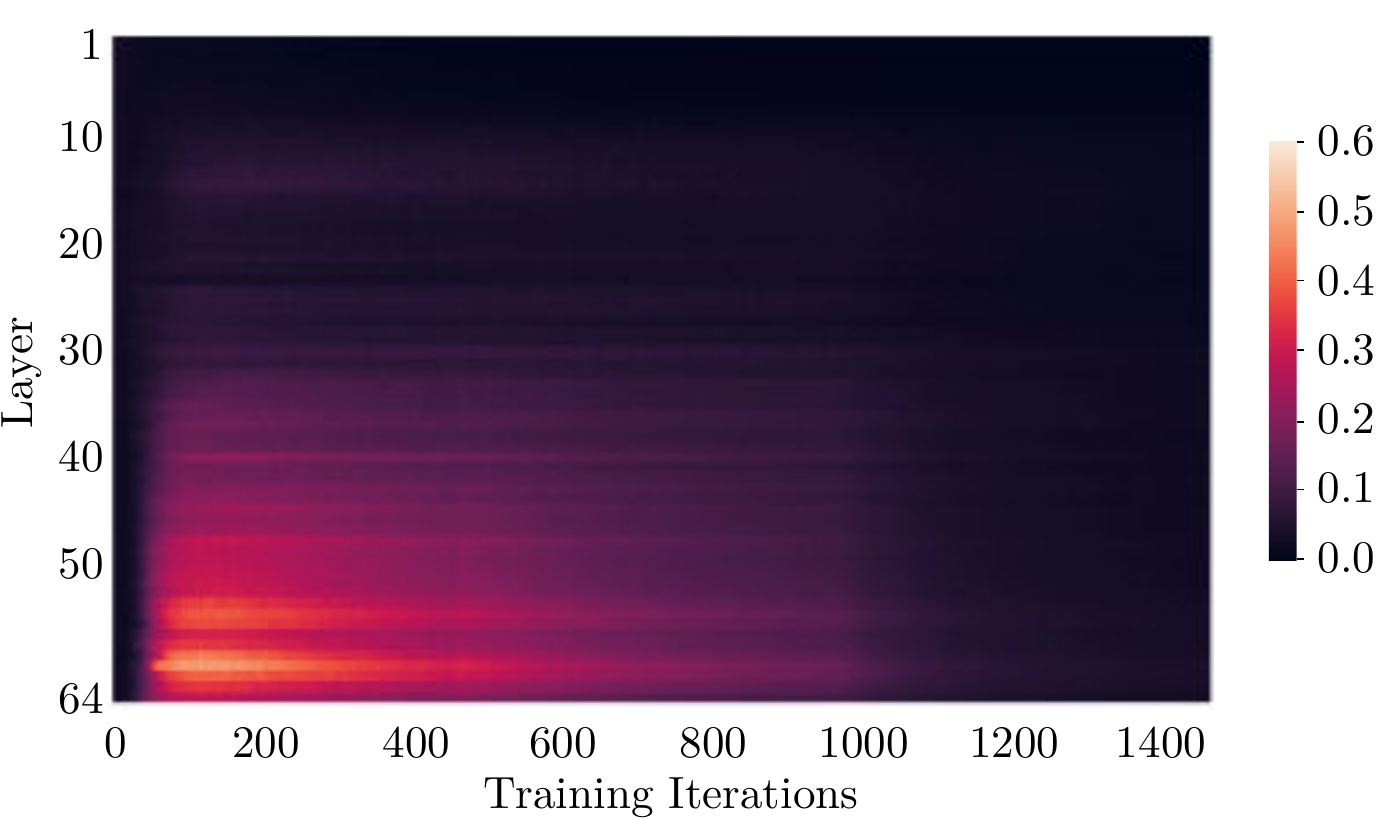}
  \end{center}
  \caption{Heat map for residual weight $|\alpha_i|$ evolution during training for 64L ReZero Transformer.\label{fig:resweight}}
\end{wrapfigure}

To probe deeper into our model, we examine the behavior of residual weights $\alpha_i$ during training for our 12 and 64 layer ReZero Transformers. The results for the 12 and 64 layer Transformer are qualitatively similar, and we show the  64 layer result in Figure \ref{fig:resweight}.
It is useful to view $|\alpha_i|$ as the amount of contribution each layer provides to the overall signal of the network.
We see that an interesting pattern emerges:
During the early iterations of training, the residual weights quickly increase to a peak value, then slowly decay to a small value later in training.
Early in training, the higher layers tend to be dominant (they peak earlier) and towards the end of training each layer is used to a similar degree.
The average $|\alpha_i|$ at the end of training is $0.0898$ and $0.0185$ for the 12 and 64 layer models respectively, which is approximately $1 / L$.
Interestingly, this pattern also occurs in the 12 layer ReZero Transformer when we initialized $\alpha$ to 1. The difference is that the model spends the first $\approx 50$ iterations forcing the $\alpha$'s to small values, before reaching a similar pattern to that in Figure \ref{fig:resweight}.
This empirical finding supports our proposal that we should initialize $\alpha=0$ even for shallow models.

\section{Conclusion}
We introduced ReZero, a simple architectural modification that facilitates signal propagation in deep networks and helps the network maintain dynamical isometry. Applying ReZero to various residual architectures -- fully connected networks, Transformers and ResNets -- we observed significantly improved convergence speeds. Furthermore, we were able to efficiently train Transformers with hundreds of layers, which has been difficult with the original architecture. We believe deeper Transformers will open the door to future exploration.

While training models with ReZero Transformers, we discovered interesting patterns in the values of residual weights of each layer $|\alpha_i|$ over the course of training. These patterns may hint towards some form of curriculum learning and allow for progressive stacking of layers to further accelerate training \citep{DBLP:conf/icml/GongHLQWL19}. Patterns of residual weights can be crucial to understand the training dynamics of such deeper networks and might be important to model performance, which we will explore in future work. 

\section*{Broader Impact}
Recent work \cite{gpt3} has shown that increasing model capacity in terms of parameter count has a substantial impact on improving model performance.
However, this increase in model capacity has also led to much longer training times and requires large GPU clusters in order to run experiments, which indirectly contributes to the carbon footprints generated from model training.
In addition, \citep{DBLP:conf/acl/StrubellGM19} showed that along with carbon emission cost, training deep models draws significant economic costs.
This makes it more difficult for less well-funded research labs and start-ups to effectively train state-of-the-art models.
Our work enables faster convergence without trading off model performance and is in line with recent efforts \citep{DBLP:journals/corr/abs-1907-10597} to make models more environment-friendly to train.

\bibliographystyle{unsrt}
\bibliography{refs}

\begin{thebibliography}{10}

\bibitem{lecun2015deep}
Yann LeCun, Yoshua Bengio, and Geoffrey Hinton.
\newblock Deep learning.
\newblock {\em Nature}, 521(7553), 2015.

\bibitem{he2016deep}
Kaiming He, Xiangyu Zhang, Shaoqing Ren, and Jian Sun.
\newblock Deep residual learning for image recognition.
\newblock In {\em CVPR}, 2016.

\bibitem{klambauer2017self}
G{\"u}nter Klambauer, Thomas Unterthiner, Andreas Mayr, and Sepp Hochreiter.
\newblock Self-normalizing neural networks.
\newblock In {\em NIPS}, 2017.

\bibitem{radford2019language}
Alec Radford, Jeffrey Wu, Rewon Child, David Luan, Dario Amodei, and Ilya
  Sutskever.
\newblock Language models are unsupervised multitask learners.
\newblock {\em OpenAI Blog}, 1(8), 2019.

\bibitem{poole2016exponential}
Ben Poole, Subhaneil Lahiri, Maithra Raghu, Jascha Sohl-Dickstein, and Surya
  Ganguli.
\newblock Exponential expressivity in deep neural networks through transient
  chaos.
\newblock In {\em NIPS}, pages 3360--3368, 2016.

\bibitem{he2015delving}
Kaiming He, Xiangyu Zhang, Shaoqing Ren, and Jian Sun.
\newblock Delving deep into rectifiers: Surpassing human-level performance on
  imagenet classification.
\newblock In {\em ICCV}, 2015.

\bibitem{srivastava2015highway}
Rupesh~Kumar Srivastava, Klaus Greff, and J{\"u}rgen Schmidhuber.
\newblock Highway networks.
\newblock {\em arXiv preprint arXiv:1505.00387}, 2015.

\bibitem{mishkin2015all}
Dmytro Mishkin and Jiri Matas.
\newblock All you need is a good init.
\newblock {\em arXiv preprint arXiv:1511.06422}, 2015.

\bibitem{xiao2018dynamical}
Lechao Xiao, Yasaman Bahri, Jascha Sohl-Dickstein, Samuel~S Schoenholz, and
  Jeffrey Pennington.
\newblock Dynamical isometry and a mean field theory of cnns: How to train
  10,000-layer vanilla convolutional neural networks.
\newblock {\em arXiv preprint arXiv:1806.05393}, 2018.

\bibitem{zhang2019fixup}
Hongyi Zhang, Yann~N Dauphin, and Tengyu Ma.
\newblock Fixup initialization: Residual learning without normalization.
\newblock {\em arXiv preprint arXiv:1901.09321}, 2019.

\bibitem{ioffe2015batch}
Sergey Ioffe and Christian Szegedy.
\newblock Batch normalization: Accelerating deep network training by reducing
  internal covariate shift.
\newblock {\em arXiv preprint arXiv:1502.03167}, 2015.

\bibitem{ba2016layer}
Jimmy~Lei Ba, Jamie~Ryan Kiros, and Geoffrey~E Hinton.
\newblock Layer normalization.
\newblock {\em arXiv preprint arXiv:1607.06450}, 2016.

\bibitem{pennington2017resurrecting}
Jeffrey Pennington, Samuel Schoenholz, and Surya Ganguli.
\newblock Resurrecting the sigmoid in deep learning through dynamical isometry:
  theory and practice.
\newblock In {\em NIPS}, 2017.

\bibitem{al2019character}
Rami Al-Rfou, Dokook Choe, Noah Constant, Mandy Guo, and Llion Jones.
\newblock Character-level language modeling with deeper self-attention.
\newblock In {\em AAAI}, volume~33, 2019.

\bibitem{montufar2014number}
Guido~F Montufar, Razvan Pascanu, Kyunghyun Cho, and Yoshua Bengio.
\newblock On the number of linear regions of deep neural networks.
\newblock In {\em NIPS}, 2014.

\bibitem{pennington2018emergence}
Jeffrey Pennington, Samuel~S Schoenholz, and Surya Ganguli.
\newblock The emergence of spectral universality in deep networks.
\newblock {\em arXiv preprint arXiv:1802.09979}, 2018.

\bibitem{glorot2010understanding}
Xavier Glorot and Yoshua Bengio.
\newblock Understanding the difficulty of training deep feedforward neural
  networks.
\newblock In {\em NIPS}, 2010.

\bibitem{saxe2013exact}
Andrew~M Saxe, James~L McClelland, and Surya Ganguli.
\newblock Exact solutions to the nonlinear dynamics of learning in deep linear
  neural networks.
\newblock {\em arXiv preprint arXiv:1312.6120}, 2013.

\bibitem{schoenholz2016deep}
Samuel~S Schoenholz, Justin Gilmer, Surya Ganguli, and Jascha Sohl-Dickstein.
\newblock Deep information propagation.
\newblock {\em arXiv preprint arXiv:1611.01232}, 2016.

\bibitem{yang2017mean}
Ge~Yang and Samuel Schoenholz.
\newblock Mean field residual networks: On the edge of chaos.
\newblock In {\em NIPS}, 2017.

\bibitem{gilboa2019dynamical}
Dar Gilboa, Bo~Chang, Minmin Chen, Greg Yang, Samuel~S Schoenholz, Ed~H Chi,
  and Jeffrey Pennington.
\newblock Dynamical isometry and a mean field theory of lstms and grus.
\newblock {\em arXiv preprint arXiv:1901.08987}, 2019.

\bibitem{he2016identity}
Kaiming He, Xiangyu Zhang, Shaoqing Ren, and Jian Sun.
\newblock Identity mappings in deep residual networks.
\newblock In {\em European conference on computer vision}. Springer, 2016.

\bibitem{goyal2017accurate}
Priya Goyal, Piotr Doll{\'a}r, Ross Girshick, Pieter Noordhuis, Lukasz
  Wesolowski, Aapo Kyrola, Andrew Tulloch, Yangqing Jia, and Kaiming He.
\newblock Accurate, large minibatch sgd: Training imagenet in 1 hour.
\newblock {\em arXiv preprint arXiv:1706.02677}, 2017.

\bibitem{hardt2016identity}
Moritz Hardt and Tengyu Ma.
\newblock Identity matters in deep learning.
\newblock {\em arXiv preprint arXiv:1611.04231}, 2016.

\bibitem{he2019bag}
Tong He, Zhi Zhang, Hang Zhang, Zhongyue Zhang, Junyuan Xie, and Mu~Li.
\newblock Bag of tricks for image classification with convolutional neural
  networks.
\newblock In {\em CVPR}, 2019.

\bibitem{de2020batch}
Soham De and Samuel~L Smith.
\newblock Batch normalization biases deep residual networks towards shallow
  paths.
\newblock {\em arXiv preprint arXiv:2002.10444}, 2020.

\bibitem{vaswani2017attention}
Ashish Vaswani, Noam Shazeer, Niki Parmar, Jakob Uszkoreit, Llion Jones,
  Aidan~N Gomez, {\L}ukasz Kaiser, and Illia Polosukhin.
\newblock Attention is all you need.
\newblock In {\em NIPS}, 2017.

\bibitem{duchi2011adaptive}
John Duchi, Elad Hazan, and Yoram Singer.
\newblock Adaptive subgradient methods for online learning and stochastic
  optimization.
\newblock {\em Journal of machine learning research}, 12(Jul), 2011.

\bibitem{savarese2016learning}
Pedro~HP Savarese, Leonardo~O Mazza, and Daniel~R Figueiredo.
\newblock Learning identity mappings with residual gates.
\newblock {\em arXiv preprint arXiv:1611.01260}, 2016.

\bibitem{smith2019super}
Leslie~N Smith and Nicholay Topin.
\newblock Super-convergence: Very fast training of neural networks using large
  learning rates.
\newblock In {\em Artificial Intelligence and Machine Learning for Multi-Domain
  Operations Applications}, volume 11006, page 1100612. International Society
  for Optics and Photonics, 2019.

\bibitem{DBLP:conf/naacl/DevlinCLT19}
Jacob Devlin, Ming{-}Wei Chang, Kenton Lee, and Kristina Toutanova.
\newblock {BERT:} pre-training of deep bidirectional transformers for language
  understanding.
\newblock In Jill Burstein, Christy Doran, and Thamar Solorio, editors, {\em
  NAACL}, 2019.

\bibitem{DBLP:conf/iclr/MerityX0S17}
Stephen Merity, Caiming Xiong, James Bradbury, and Richard Socher.
\newblock Pointer sentinel mixture models.
\newblock In {\em ICLR}, 2017.

\bibitem{mahoney2009}
Matt Mahoney.
\newblock Large text compression benchmark, 2009.

\bibitem{DBLP:journals/corr/abs-1910-05895}
Toan~Q. Nguyen and Julian Salazar.
\newblock Transformers without tears: Improving the normalization of
  self-attention.
\newblock {\em CoRR}, abs/1910.05895, 2019.

\bibitem{DBLP:journals/corr/abs-1910-10683}
Colin Raffel, Noam Shazeer, Adam Roberts, Katherine Lee, Sharan Narang, Michael
  Matena, Yanqi Zhou, Wei Li, and Peter~J. Liu.
\newblock Exploring the limits of transfer learning with a unified text-to-text
  transformer.
\newblock {\em CoRR}, abs/1910.10683, 2019.

\bibitem{xiong2020layer}
Ruibin Xiong, Yunchang Yang, Di~He, Kai Zheng, Shuxin Zheng, Chen Xing,
  Huishuai Zhang, Yanyan Lan, Liwei Wang, and Tie-Yan Liu.
\newblock On layer normalization in the transformer architecture.
\newblock {\em arXiv preprint arXiv:2002.04745}, 2020.

\bibitem{tnlg2020}
Microsoft.
\newblock Turing-nlg: A 17-billion-parameter language model, 2020.

\bibitem{DBLP:conf/icml/GongHLQWL19}
Linyuan Gong, Di~He, Zhuohan Li, Tao Qin, Liwei Wang, and Tie{-}Yan Liu.
\newblock Efficient training of {BERT} by progressively stacking.
\newblock In Kamalika Chaudhuri and Ruslan Salakhutdinov, editors, {\em ICML},
  volume~97 of {\em Proceedings of Machine Learning Research}, 2019.

\bibitem{gpt3}
Tom~B. Brown, Benjamin Mann, Nick Ryder, Melanie Subbiah, Jared Kaplan,
  Prafulla Dhariwal, Arvind Neelakantan, Pranav Shyam, Girish Sastry, Amanda
  Askell, Sandhini Agarwal, Ariel Herbert-Voss, Gretchen Krueger, Tom Henighan,
  Rewon Child, Aditya Ramesh, Daniel~M. Ziegler, Jeffrey Wu, Clemens Winter,
  Christopher Hesse, Mark Chen, Eric Sigler, Mateusz Litwin, Scott Gray,
  Benjamin Chess, Jack Clark, Christopher Berner, Sam McCandlish, Alec Radford,
  Ilya Sutskever, and Dario Amodei.
\newblock Language models are few-shot learners.
\newblock 2020.

\bibitem{DBLP:conf/acl/StrubellGM19}
Emma Strubell, Ananya Ganesh, and Andrew McCallum.
\newblock Energy and policy considerations for deep learning in {NLP}.
\newblock In Anna Korhonen, David~R. Traum, and Llu{\'{\i}}s M{\`{a}}rquez,
  editors, {\em Proceedings of the 57th Conference of the Association for
  Computational Linguistics, {ACL} 2019, Florence, Italy, July 28- August 2,
  2019, Volume 1: Long Papers}. ACL, 2019.

\bibitem{DBLP:journals/corr/abs-1907-10597}
Roy Schwartz, Jesse Dodge, Noah~A. Smith, and Oren Etzioni.
\newblock Green {AI}.
\newblock {\em CoRR}, abs/1907.10597, 2019.

\bibitem{DBLP:journals/corr/HendrycksG16}
Dan Hendrycks and Kevin Gimpel.
\newblock Bridging nonlinearities and stochastic regularizers with gaussian
  error linear units.
\newblock {\em CoRR}, abs/1606.08415, 2016.

\bibitem{DBLP:journals/corr/abs-1904-00962}
Yang You, Jing Li, Jonathan Hseu, Xiaodan Song, James Demmel, and Cho{-}Jui
  Hsieh.
\newblock Reducing {BERT} pre-training time from 3 days to 76 minutes.
\newblock {\em CoRR}, abs/1904.00962, 2019.

\bibitem{hochreiter1997long}
Sepp Hochreiter and J{\"u}rgen Schmidhuber.
\newblock Long short-term memory.
\newblock {\em Neural computation}, 9(8), 1997.

\end{thebibliography}

\appendix

\section{Vanishing singular values in Transformers}
\label{sec:singval}
Two crucial components relevant to the signal propagation in the Transformer include LayerNorm \citep{ba2016layer} and (multi-head) self attention \citep{vaswani2017attention}.
In this section, we argue that neither component by itself or in conjunction with a vanilla residual connection can satisfy dynamical isometry for all input signals. 

\subsection{LayerNorm}
Layer normalization removes the mean and scales the variance  over all neurons of a given layer  and introduces learnable parameters  $\gamma$ and $\beta$ to re-scale the variance and shift the mean according to
\be
\text{LayerNorm}(\bs x)= \frac{\bs x -\text{E}(\bs x)}{\sqrt{\text{Var}(x)}}\times \gamma+\beta\,.\label{layernorm}
\ee
It is clear from this definition that perturbing an input $x$ by a transformation that purely shifts either its mean or variance will leave the output unchanged. These perturbations, therefore, give rise to two vanishing singular values of the input-output Jacobian. In the Transformer architecture \citep{vaswani2017attention}, the norm is applied to each of the $n$ elements of the input sentence, leading to a total of $2\times n$ vanishing singular values of the Jacobian for each Transformer layer.

\subsection{Self-Attention}
Self-attention allows the model to relate content located across different positions by computing a weighted sum of an input sequence. Specifically, the $n\times d$ matrix $\bs x$ contains an input sequence of $n$ rows containing $d$-dimensional embedding vectors, from which we can evaluate the query, key and value matrices $\bs Q,\bs K,\bs V = \bs x\cdot \bs W^{Q,K,V}$, where the $W^{Q,K,V}$ matrices are $d \times d$.
The scaled dot-product attention then is given by
\be
\text{Attention}(\bs Q,\bs K,\bs V)=\text{softmax}\left(\bs Q\cdot\bs K^\top / \sqrt{d}\right)\cdot\bs V\,.
\ee
In general, the singular value spectrum of the Jacobian of this attention process is complicated. Rather than studying it in full generality, we now merely argue that for some inputs $\bs x$ and weights $\bs W^{Q,K,V}$  the Jacobian has a large number of vanishing singular values (a claim we evaluate empirically in the paper).
Consider weights or inputs such that each of the arguments of the softmax function is small compared to 1. The softmax function then simply returns a $n\times n$ dimensional matrix filled with entries that all approximate $1/n$. This means that the attention function projects all embedding vectors of the input sequence onto a single diagonal direction. This implies that out of the $n\times d$ Jacobian singular values only $d$ are non-vanishing and hence much of the input signal is lost.
A residual connection can restore some of the lost signals, but even then some perturbations are amplified while others are attenuated. 
This example demonstrates that self-attention is incompatible with dynamical isometry and unimpeded signal propagation in deep Transformer networks.
It is easy to verify that the same conclusion holds for multi-head attention. A careful initialization of the weights might alleviate some of these issues, but we are not aware of any initialization scheme that would render a Transformer layer consistent with dynamical isometry.

\section{Convergence speed experimental hyperparameters}
\label{sec:appendix-tx1}
For all model variants in Section 6.2, we control the batch size to be 1080, number of layers to 12, feed-forward and attention dropout to 20\%, hidden and embedding size to 512 units, context length to 512, the attention heads to 2, and GELU \citep{DBLP:journals/corr/HendrycksG16} activation in the point-wise feed-forward layer. To accommodate large batch training we use the LAMB optimizer \citep{DBLP:journals/corr/abs-1904-00962} with a fixed learning rate of $0.016$. Although learning rate schedules tend to improve performance \citep{DBLP:conf/naacl/DevlinCLT19}, we omit them to simplify our training process.
Training is performed on 8x V100 GPUs for at most a few days.

\section{Deep Transformers experimental hyperparameters}
\label{sec:appendix-tx2}
In Section 6.3, in order to examine whether our approach scales to deeper Transformers, we scale our 12 layer ReZero Transformer from Section 6.2 to 64 layers and 128 layers and compare it against the vanilla Transformer (\textit{Post-Norm}). Due to memory constraints, we decreased the hidden size from 512 to 256 and reduced batch size to 304 and 144 for the 64 layer and 128 layer model respectively. Following guidelines from \citep{DBLP:journals/corr/abs-1904-00962} we also adjusted the learning rate according to $0.0005 \times \sqrt{\text{batch size}}$. For all models in our experiments we limit training to a maximum of 100 training epochs.
Training is performed on 8x V100 GPUs for at most a few days.

\section{Review of residual gates for deep signal propagation}
\label{sec:residualgates}
In this section we give a chronological but light review of residual gates that are designed to preserve signals as they propagate deep into neural networks.

\subsection{Highway Networks}
Highway Networks \cite{srivastava2015highway}, based on ideas from LSTM \cite{hochreiter1997long}, were the first feedforward neural networks with hundreds of layers. This architecture employs gating units that learn to regulate signal flow through the network. Specifically, the authors define transform and carry gates $T[ \bs W_T](\bs x)$ and $C[ \bs W_C](\bs x)$, with weights $\bs W_{T,C}$ that act explicitly non-linearly on the signal $\bs x$. When combined with some block $F(\bs x_i)$ of a deep network, this gives the transformation
\be\label{highway}
\bs x_{i+1} = C[ \bs W_C](\bs x) \cdot \bs x_i + T[ \bs W_T](\bs x)\cdot  F(\bs x_i)\,.
\ee
The authors further propose to simplify the architecture according to $C\equiv 1-T$, and using a simple Transform gate of the form $T[ \bs W_T](\bs x) \equiv \sigma(\bs W_T^\top \cdot \bs x +\bs b_T )$, where $\sigma$ denotes some activation function. The bias is initialized to be negative, as to bias the network towards carry behavior, e.g., $C$, but the network is not initialized as the identity map.
    
The proposal of Highway Networks lead to Gated ResNet \cite{savarese2016learning}, in which there exists a single additional parameter that parametrizes the gates as $ \bs W_T = \bs 0$, $\bs b_T = \alpha$, $C = 1 - T$.
    
Any feed-forward network could be written in the form (13), and ReZero corresponds to the simple choice $ \bs W_T = \bs W_C = \bs 0$, $\bs b_T = \alpha$, $\bs b_C = 1$. In contrast to Highway Networks, in ReZero the gates are independent of the input signal. We compare the performance of Gated ResNets to ReZero ResNets in Section 5.

\subsection{ResNets}
ResNets \cite{he2016deep} introduced the simple residual connection 
\be\label{resnetconnection}
\bs x_{i+1} = \sigma \left(\bs x_i +   F(\bs x_i) \right)\,,
\ee
that has been extremely successful in training deep networks. Just as Highway Networks, these residual connections are not initialized to give the identity map.
    
\subsection{Pre-activation ResNets}
Soon after the introduction of ResNets it was realized in \cite{he2016identity} that applying the activation function $\sigma$ prior to the residual connection allows for better performance. Schematically, we have the pre-activation connection
\be\label{preactresnetconnection}
\bs x_{i+1} =\bs x_i +    F(\bs x_i)\,,
\ee
where we absorbed the activation function into the block $F(\bs x_i)$. This finding of improved performance is consistent with improved signal propagation, since the residual connection is not modulated by the activation function.

\subsection{Zero $\gamma$}
Residual networks often contain normalization layers in which the signal is rescaled by learnable parameters $\gamma$ which is referred to the Zero $\gamma$ \cite{goyal2017accurate,hardt2016identity,he2019bag}. If the last element before a residual connection happens to be a normalization layer, then initializing these $\gamma$ to zero has been found to improve convergence speed and accuracy. This method is in spirit very similar to the ReZero architecture. However, it potentially zero-initializes many parameters for each block, and is only applicable when a normalization layer exists.
    
\subsection{FixUp}
FixUp initialization \cite{zhang2019fixup} carefully rescales the initialization scheme in order to avoid vanishing or exploding gradients, without the use of normalization techniques. In particular, this scheme is implemented via the following Rules (verbatim from \cite{zhang2019fixup}):
    \begin{enumerate}
        \item Initialize the classification layer and the last layer of each residual branch to 0.
        \item Initialize every other layer using a standard method (e.g., He et al. \cite{he2015delving}), and scale only the  weight layers inside residual branches by $L^{-1/(2m-2)}$.
        \item Add a scalar multiplier (initialized at 1) in every branch and a scalar bias (initialized at $0$) before each convolution, linear, and element-wise activation layer.
    \end{enumerate}
The authors emphasize that Rule 2 is the essential part. ReZero is simpler and similar to the first part of Rule 3, but the initialization differs.

\subsection{SkipInit}
\cite{de2020batch} proposes to replace BatchNorm layers with a single scalar initialized at a small value. SkipInit is only applicable when normalization layers exist.

\subsection{ReZero}
ReZero is the simplest iteration achieving the goal of deep signal propagation. Schematically, the ReZero architecture is
\be
\bs x_{i+1} = \bs x_i + \alpha_i F(\bs x_i)\,.
\ee
The rule to implement ReZero is
\begin{enumerate}
    \item For every block add a scalar multiplier $\alpha$ (initialized at 0) and a residual connection.
\end{enumerate}

\section{CIFAR-10 experiments}\label{sec:cifarhyperparameters}
In this section we briefly describe the hyperparameters used in the image recognition experiments performed in \S5. For all these experiments we used PyTorch version 1.2.0 (we have observed some inconsistencies in results with other PyTorch versions that may be due to different default initializations).
CIFAR10 experiments tend to take less than an hour on a single RTX 2080TI GPU.

\subsection{Step-down schedule}
In Table 1 we compare the inference accuracy of different network architectures after training with identical hyperparameters a learning-rate schedule that decreases in three steps, as in \cite{he2016deep}. The initial learning rate is $0.1$ and decreases by a factor of $10$ at $100$ and $150$ epochs. The models are trained for a total of $200$ epochs. We use the SGD optimizer with a batch size of $128$, weight decay of $5\times 10^{-4}$ and momentum $0.9$.

\subsection{Superconvergence schedule}
To demonstrate superconvergence we use a one-cycle learning rate schedule, as described in \cite{smith2019super} and closely follow the setup by Fast AI referenced in the text. In particular, the learning rate of the SGD optimizer evolves as follows over $45$ epochs. The initial learning rate is $0.032$ and linearly increases with each iteration to reach $1.2$ after  $10\%$ of the total number of iterations has been reached. Then, the learning rate linearly decreases to return to $0.032$ when $90\%$ of the total steps. Thereafter, the learning rate linearly decays to a final value of $0.001$ at the end of training. The SGD momentum varies between $0.85$ and $0.95$, mirroring the triangular learning rate, as is standard for the one-cycle policy in this setup \cite{smith2019super}. Weight decay is set to $2\times 10^{-4}$ and the batch size is $512$.

The residual weights cannot tolerate the extremely large learning rates required for the super-convergence phenomenon. For this reason we keep the learning rate of the residual weights at $0.1$ throughout training.

\section{Datasets}
We note that for all our experiments, we follow the official training, validation and test splits of the respective datasets.

\end{document}